\documentclass[11pt,a4paper]{article}
\usepackage[hyperref]{emnlp2020}
\usepackage{times}
\usepackage{latexsym}
\usepackage{booktabs}

\usepackage{microtype}
\usepackage{enumitem}

\aclfinalcopy 

\title{PySBD: Pragmatic Sentence Boundary Disambiguation}

\author{Nipun Sadvilkar \\
  Episource LLC \\
  \texttt{nipun.sadvilkar@episource.com} \\\And
  Mark Neumann \\
  Allen Institute for Artificial Intelligence \\
  \texttt{markn@allenai.org} \\}

\date{}

\begin{document}
\maketitle
\begin{abstract}
In this paper, we present a rule-based sentence boundary disambiguation Python package that works out-of-the-box for 22 languages. We aim to provide a realistic segmenter which can provide logical sentences even when the format and domain of the input text is unknown. In our work, we adapt the \emph{Golden Rules Set} (a language specific set of sentence boundary exemplars) originally implemented as a ruby gem \emph{pragmatic\_segmenter}\footnote{\url{https://github.com/diasks2/pragmatic_segmenter}} which we ported to Python with additional improvements and functionality. PySBD passes 97.92\% of the Golden Rule Set examplars for English, an improvement of 25\% over the next best open source Python tool.
\end{abstract}

\section{Introduction}
Sentence Boundary Disambiguation (SBD), also known as sentence boundary detection, is a key underlying task for natural language processing. In many NLP pipelines, gold standard SBD often assumed, and acts as a primary input to downstream NLP tasks such as machine translation, named entity recognition and coreference resolution. However, in real world scenarios, text occurs in a variety of input modalities, such as HTML forms, PDFs and word processing doccument formats.

Although SBD is considered to be a simple problem, it becomes more complex in other domains due to unorthodox use of punctuation symbols. For example, drug names in medical documents, case citations in legal text and references in academic articles all use punctuation in ways which are uncommon in newswire documents. Simple SBD approaches for English web text (treating - “?!:;.” - as end of sentence markers) covers a majority of cases, but an ideal SBD system should be able to disambiguate these edge case scenarios and be robust to \textit{known} textual variation.

Our contributions in this paper are describing an open-source, freely available tool for pragmatic sentence boundary disambiguation. In particular, we describe the implementation details of PySBD, evaluate it in comparison to other open source SBD tools and discuss it's natural extensibility due to it's rule based nature.

\section{Related Work}
Sentence segmentation methods can be broadly divided into 3 approaches: i) rules-based, requiring hand crafted rules/heuristics; ii) Supervised machine learning, requiring annotated datasets, and iii) Unsupervised machine learning, requiring distributional statistics derived from raw text. 

\cite{palmer-hearst-1997-adaptive} use decision trees and neural networks in a supervised, feature based SBD model, requiring part of speech information and training data. \cite{kiss-strunk-2006-unsupervised} design punkt, an unsupervised SBD model centered around the observation that abbreviations are the main confounders for rule based sentence boundary models. Although it is unsupervised, punkt requires the computation of various co-occurence and distributional statistics of a relevant corpora; as PySBD is rule-based, it does not require a initial corpus of text. \cite{evang-etal-2013-elephant} cast SBD as a character sequence labelling problem and use features from a recurrent neural network language model to train a CRF labelling model.

Many SBD papers reject rule-based approaches due to non-robustness, maintainability and performance. We reject these conclusions, and instead focus on the \textit{positive} features of rule-based systems - namely that their errors are interpretable, rules can be adjusted incrementally and their performance is often on-par with learnt statistical models.

\paragraph{\textbf{Issues with benchmarks on PTB/WSJ corpora}}

SBD systems have historically been benchmarked on the Wall Street Journal/Penn Treebank corpora \citep{read-etal-2012-sentence}. The majority of the sentences found in the Penn Tree Bank are sentences that end with a regular word followed by a period, testing the same sentence boundary cases repeatedly. In the Brown Corpus 90\% of potential sentence boundaries come after a regular word. Although the Wall Street Journal corpus is richer with numerical values, abbreviations and only 53\% according to \citep{gale-church-1993-program} of sentences end with a regular word followed by a period \citep{mikheev-2002-periods}. 

Given that commonly used training/evaluation corpora do not contain a particularly large amount of sentence marker variation, we use a \emph{Golden Rule Set} to enumerate edge cases observed in sentence boundaries. The \emph{Golden Rule Set} contains 48 hand-constructed rules, designed to cover sentence boundaries across a variety of domains. The GRS is interpretable (each rule targets a specific type of sentence boundary) and easy to extend with new examples of particular sentence boundary markers.

\section{Implementation}

PySBD is divided into four high level components: The Segmenter, Processor, Language and Cleaner sub-modules.

The \emph{Segmenter} class is the public API to PySBD. It allows a user to set up a \emph{Segmenter} in their language of choice, as well as specify additional options such as text cleaning and char\_span functionality. The \emph{Segmenter}  requires a two character ISO 639-1 code\footnote{\url{https://en.wikipedia.org/wiki/List_of_ISO_639-1_codes}} to process input text. Text extracted from a PDF or obtained from OCR systems typically contains unusually formatted text, such as line breaks in the middle of sentences. This can be handled with the \emph{doc\_type} option, or for more aggressive text cleaning, the \emph{clean} functionality performs additional pre-filtering of the input text, removing repeated and unnecessary punctuation.

The \emph{Processor} contains the sentence segmentation logic, using rules to segment the input text. The \emph{Processor} contains several groups of sentence segmentation rules, some of which are universal across languages, and some of which are language specific. These are grouped as follows: 
\begin{itemize}[noitemsep]
  \item Common
  \item Standard
  \item ListItemReplacer
  \item AbbreviationReplacer
  \item ExclamationWords
  \item BetweenPunctuation
\end{itemize}

The \emph{Processor} identifies sentence boundaries by manipulating input text in 3 stages. Firstly, rules are applied to alter the input text by adding intermediate unicode characters as placeholders to signify that particular pieces of punctuation are not sentence boundaries. The segment stage identifies true sentence boundaries by bypassing unicode characters and segments text into sentences using a much simpler regex rule. Finally, the manipulated text is transformed into original text form by replacing the unicode placeholders with their original characters.

The \emph{Language} holds all the languages supported by PySBD. Each language is built on top of two sub-components - \emph{Common} and \emph{Standard} - involving basic rules prevalent across languages. \emph{Common} rules encompass the main sentence boundary regexes; AM-PM regexes handle numerically expressed time periods; number regexes handle period/newline characters before or after single/multi-digit numbers and additional rules handle quotation, parenthesis, and numerical references within the input text. The \emph{Standard} rule set contains regex patterns to handle single/double punctuation, geolocation references, fileformat mentions and ellipsis in input text. The \emph{ListItemReplacer} rule set handles itemized, ordered/unordered lists; the \emph{AbbreviationReplacer} contains language specific common abbreviations. Finally, the \emph{ExclaimationWords} and \emph{BetweenPunctuation} rules handle language specific exclamations and more complicated punctuation cases.

In practice, text encountered in the wild is noisy, containing extraneous line breaks, unicode characters, uncommon spacing and hangovers from document structure. In order to handle this, PySBD provides an optional  component to handle such texts. The \emph{Cleaner} is passed as an option through the top-level \emph{Segmenter} component and provides text cleaning rules for cases like irregular newline characters, tables of contents, URLs, HTML tags and text involving no space between sentences. As the text cleaning rules perform a destructive operation, this feature is incompatible with the \emph{char\_span} functionality, as mapping back to character indices within the original text is no longer possible.

\section{Experimental Setup}

\paragraph{Data}
Contrary to the WSJ, Brown and GENIA datasets mentioned in \citep{read-etal-2012-sentence}, we use language specific \emph{Golden Rules Sets} for our experiments. There are total 22 Golden Rules sets  for following languages - English, Marathi, Hindi, Bulgarian, Español (Spanish), Russian, Arabic, Amharic, Armenian, Persian, Urdu, Polish, Chinese, Dutch, Danish, French, Italian, Greek, Burmese, Japanese, Deutsch (Germen), Kazakh.

The \emph{Golden Rules Sets} are devised by considering possible sentence boundaries per language as well as considering different domains. For example the English language GRS is comprised of 48 golden rules \footnote{\href{https://s3.amazonaws.com/tm-town-nlp-resources/golden_rules.txt}{Golden Rules from pragmatic\_segmenter}} from formal and informal domains to cover a wide variety of phenomena. For example, news articles are grammatically and punctually correct; scientific literature often involves numbers, abbreviations and bibliography references, and Informal domain like Web Text - E-mail, Social media text involves irregular punctuation and ellipses. To ensure our rules-based system built with respect to the \emph{Golden Rules Set} generalizes well in the real world, we have also performed a benchmark comparison on the GENIA corpus \cite{Kim2003GENIAC}, a dataset of linguistic annotations on top of the abstracts of biomedical papers. The GENIA corpus provides both raw and segmented abstracts, which we use as natural data for our evaluation.

\paragraph{Setup}
We evaluate PySBD and other segmentation tools on two corpora - the \emph{English Golden Rules Set} and GENIA corpus.

\section{Comparison to alternatives}

Table \ref{tab:grs-table} summarizes accuracy of PySBD and the alternative Python SBD modules on \emph{English Golden Rules Set} and the GENIA corpus. The supervised machine learning based sentence segmenters, stanza \cite{Qi2020StanzaAP} and spacy dependency parsing (spacy dep) \cite{spacy2} are slower compared to other python modules and seem to segment incorrectly when text contains mixed case words or abrupt punctuation within words, which is prevalent in biomedical domain. Inability to generalise on out of domain corpora is a main drawback of using supervised learning for SBD.

NLTK's \cite{Bird2006NLTKTN} \emph{PunktSentenceTokenizer} is based on an unsupervised algorithm \cite{kiss-strunk-2006-unsupervised} and fails to segment text containing brackets, itemized text and abbreviations as sentence boundaries. In contrast, practically all the modules following a rules-based approach -  blingfire \cite{Bling2020}, syntok \cite{Leitner2020} and PySBD - appear to be faster and more accurate on both corpora. Blingfire and syntok modules struggle when text has decimal numbers, abbreviations, brackets and mixed cased words prior to or following the true sentence boundary. Lastly, the 3\% drop in PySBD's accuracy on the GENIA corpus is caused by splitting itemized text into segments, whereas the GENIA abstracts contain single sentences with multiple bulleted lists. We feel that this segmentation choice comes down to preference, and both are equally valid.

Table \ref{tab:speed-table} shows the runtime performance of each module on the entire text of ``The Adventures of Sherlock Holmes" \footnote{\url{http://www.gutenberg.org/files/1661/1661-0.txt}}, which contains 10K lines and 100K words. The experiment was performed on Intel Core i5 processor running at 2.9 GHz clock speed. Blingfire module is the fastest since it is extremely performance oriented (at cost of reduced maintainability/extensibility). Although PySBD is slower than several alternatives, it is considerably faster than running full pipelines and is a good choice for users who require high accuracy segmentations. Our comparisons cover a variety of different implementations (C++, Cython, Pytorch), approaches (model-based, distributional, rule-based) and are representative of practical choices for an NLP practitioner.

\begin{table}
\centering
\begin{tabular}{@{}lcc@{}}
\textbf{Tool}      & \textbf{GRS} & \textbf{GENIA} \\ \toprule
blingfire & 75.00  & 86.95     \\
syntok    & 68.75  & 80.90    \\
spaCy     & 52.08  & 76.80    \\
spacy dep & 54.17  & 39.20  \\
stanza    & 72.92  & 63.40 \\
NLTK      & 56.25  & 87.95 \\  
PySBD     & 97.92  & 97.00 \\

\end{tabular}
\caption{Accuracy (\%) of PySBD compared to other open source SBD packages with respect to the English \emph{Golden Rule Set} and the GENIA corpus. }
\label{tab:grs-table}
\end{table}

\begin{table}
\centering
\begin{tabular}{@{}lc@{}}
\textbf{Tool}      & \textbf{Speed(ms)} \\ \toprule
blingfire & 85.24      \\
syntok    & 1764.11     \\
spaCy     & 1523.20     \\
spacy dep & 26850.69    \\
stanza    & 48383.46   \\
NLTK      & 780.49  \\  
PySBD     & 9483.96   \\

\end{tabular}
\caption{Speed benchmark on the entire text of ``The adventures of Sherlock Holmes" for PySBD compared to other open source SBD packages.}
\label{tab:speed-table}
\end{table}

\section{Discussion}

\subsection{Package Development}
We use Test-driven Development (TDD) whilst developing, first write a test for one of the rule from \emph{Golden Rules Set} that fails intentionally, before writing functional code to pass the test. The approach used by our python module is rules-based. We employ Python's standard library module called re\footnote{\url{https://docs.python.org/3/library/re.html}} which provides regular expression (regex) matching operations for text processing.

\subsection{Non-destructive Segmentation}
Our module does non-destructive sentence tokenization, as when dealing with noisy text, character offsets into the original documents are often desirable. The indices are obtained after postprocess stage by mapping post-processed sentence start indices into the original input text. Upon enabling this feature, the output format is a list of \emph{TextSpan} objects containing sentence, start and end character indices. The character spans makes it easy to navigate to any sentence within the unaltered original text.

\subsection{Multilingual Implementation}

NLP research predominantly focuses on developing dataset and methods for English language despite the many benefits of working on other languages \cite{ruder2020}. An advantage of PySBD's rule-based approach is straightforward extension to new languages. PySBD has support for 22 languages spanning many language families, each having its own \emph{Golden Rules Set}. Adding support for a new language involves adding language specific rules to the \textit{Golden Rules Set} and adding language specific punctuation markers to a new language module. This modular language support allows PySBD to be maintained in a community driven way by open source NLP practitioners. For example, we extended the original Ruby pragmatic segmeter by adding support for Marathi by forming the Golden Rules and identifying Marathi language specific sentence syntax, punctuations and abbreviations resulting in a usable module within 1 hour of work. 
If you would like to contribute to the PySBD module by updating existing GRS or by adding support for new language then refer to our contributing guidelines\footnote{\href{https://github.com/nipunsadvilkar/pySBD/blob/master/CONTRIBUTING.md}{PySBD Contributing Guidelines}} to get you started.
\begin{table}
\centering
\begin{tabular}{@{}lc@{}}
\textbf{Language}      & \textbf{Accuracy (\%)} \\ \toprule
Amharic   & 80.95\%  \\
Arabic    & 70.40\%  \\
Armenian  & 63.75\%  \\
Bulgarian & 93.35\%  \\
Burmese   & 48.05\%  \\
Chinese   & 85.35\%  \\
Danish    & 91.40\%  \\
Deutsch   & 80.95\%  \\
Dutch     & 91.40\%  \\
French    & 91.90\%  \\
Greek     & 91.05\%  \\
Hindi     & 88.50\%  \\
Italian   & 90.55\%  \\
Japanese  & 96.45\%  \\
Kazakh    & 63.20\%  \\
Marathi   & 92.60\%  \\
Persian   & 84.95\%  \\
Polish    & 55.48\%  \\
Russian   & 88.55\%  \\
Spanish   & 92.65\%  \\
Urdu      & 77.55\%  \\ \hline
\end{tabular}
\caption{Accuracy of PySBD's multilingual modules on the OPUS-100 multilingual corpus test sets, containing 2000 sentences per language. Each language module build with respect to its own GRS.}
\label{tab:multilang-opus-table}
\end{table}

We benchmarked PySBD on the OPUS-100 parallel multilingual corpus, covering 100 languages \cite{Tiedemann2012ParallelDT}. We used the test sets of 21 of the languages excluding English which contain 2000 sentences per language (Due to unavailability of the test set for Armenian, we used its train set, containing 7000 sentences). Due to noisy nature of OPUS (on inspection, multiple sentences were present on individual lines in the test sets) and lack of language specific knowledge to form rules and abbreviation list we observed weak performance in a few languages like Burmese, Polish, Kazakh, Armenian, etc. Shortcomings of such languages can be improved in community driven way by collaborating with multilingual NLP practitioners.

\section{Conclusion}
In this paper, we have described PySBD, a \textit{pragmatic} sentence boundary disambiguation model. PySBD is open source, has over 98\% test coverage and integrates easily with existing natural language processing pipelines. PySBD currently supports 22 languages, and is easily extensible, with 57 projects depending on it at the time of writing. Although slower than some alternatives implemented in low level languages such as C++, PySBD successfully disambiguates 97\% of sentence boundaries in a \textit{Golden Rule Set} and is robust across domains and noisy text. 

\bibliographystyle{acl_natbib}

\bibliography{pysbd}

\appendix

\end{document}